\documentclass[11pt,a4paper]{article}
\usepackage[hyperref]{acl2020}
\usepackage{times}
\usepackage{latexsym}

\usepackage{microtype}

\aclfinalcopy % Uncomment this line for the final submission
\def\aclpaperid{384} %  Enter the acl Paper ID here

\usepackage{times}
\usepackage{latexsym}
\usepackage{graphicx}
\usepackage{tabularx}
\usepackage{wrapfig}
\usepackage[normalem]{ulem}
\usepackage{amsmath}
\usepackage{subcaption}
\usepackage{xcolor}
\usepackage{multirow}
\usepackage{multicol}
\usepackage{makecell}
\usepackage{comment}
\usepackage{amssymb}
\def\squiggly{\bgroup \markoverwith{\textcolor{red}{\lower3.5\p@\hbox{\sixly \char58}}}\ULon}

\usepackage{algorithmicx}
\usepackage[ruled]{algorithm}
\usepackage{algpseudocode}
\algnewcommand{\LineComment}[1]{\State \(\triangleright\) #1}

\usepackage{url} 
\usepackage{xcolor,colortbl}

\usepackage{mathtools}

\DeclarePairedDelimiterX{\infdivx}[2]{(}{)}{%
  #1\;\delimsize\|\;#2%
}
\newcommand{\infdiv}{D_{KL}\infdivx}

\usepackage{paralist}

\definecolor{Gray}{gray}{0.8}

\title{Unsupervised Opinion Summarization with Noising and Denoising}

\author{
  Reinald Kim Amplayo \and
  Mirella Lapata \\
  Institute for Language, Cognition and Computation \\
  School of Informatics, University of Edinburgh \\
  \tt{reinald.kim@ed.ac.uk, mlap@inf.ed.ac.uk}
}

\date{}

\makeatletter
\newcommand{\thickhline}{%
    \noalign {\ifnum 0=`}\fi \hrule height 1pt
    \futurelet \reserved@a \@xhline
}
\makeatother

\begin{document}
\maketitle
\begin{abstract}
  The supervised training of high-capacity models on large datasets
  containing hundreds of thousands of document-summary pairs is
  critical to the recent success of deep learning techniques for
  abstractive summarization. Unfortunately, in most domains (other
  than news) such training data is not available and cannot be easily
  sourced. In this paper we enable the use of supervised learning for
  the setting where there are only documents available (e.g.,~product
  or business reviews) without ground truth summaries. We create a
  synthetic dataset from a corpus of user reviews by sampling a
  review, pretending it is a summary, and generating noisy versions
  thereof which we treat as pseudo-review input. We introduce several
  linguistically motivated noise generation functions and a
  summarization model which learns to denoise the input and generate
  the original review.  At test time, the model accepts genuine
  reviews and generates a summary containing salient opinions,
  treating those that do not reach consensus as noise.  Extensive
  automatic and human evaluation shows that our model brings
  substantial improvements over both abstractive and extractive
  baselines.

\end{abstract}

\section{Introduction}
\label{sec:intro}

The proliferation of massive numbers of online product, service, and
merchant reviews has provided strong impetus to develop systems that
perform opinion mining automatically \cite{pang2007opinion}.  The vast
majority of previous work \cite{hu2006opinion} breaks down the problem
of opinion aggregation and summarization into three inter-related
tasks involving aspect extraction \cite{mukherjee2012aspect},
sentiment identification \cite{pang2002thumbs,pang2004sentimental},
and summary creation based on \emph{extractive}
\cite{radev2004centroid,lu2009rated} or \emph{abstractive} methods
\cite{ganesan2010opinosis,carenini2013multi,gerani2014abstractive,difabbrizio2014hybrid}.
Although potentially more challenging, abstractive approaches seem
more appropriate for generating informative and concise summaries,
e.g.,~by performing various rewrite operations (e.g.,~deletion of
words or phrases and insertion of new ones) which go beyond simply
copying and rearranging passages from the original opinions.

Abstractive summarization has enjoyed renewed interest in recent years
thanks to the availability of large-scale datasets
\cite{nytcorpus,hermann-nips15,newsroom-naacl18,liu2018generating,fabbri-etal-2019-multi}
which have driven the development of neural architectures for
summarizing single and multiple documents. Several approaches
\cite{see2017get,asli-multiagent18,paulus2017deep,gehrmann2018bottom,liu2018generating,perez-beltrachini-etal-2019-generating,liu-lapata-2019-hierarchical,wang2016neural}
have shown promising results with sequence-to-sequence models that
encode one or several source documents and then decode the learned
representations into an abstractive summary.

The supervised training of high-capacity models on large datasets
containing hundreds of thousands of document-summary pairs is critical
to the recent success of deep learning techniques for abstractive
summarization. Unfortunately, in most domains (other than news) such
training data is not available and cannot be easily sourced.  For
instance, manually writing opinion summaries is practically
impossible since an annotator must read all available reviews for a
given product or service which can be prohibitively many. Moreover,
different types of products impose different restrictions on the
summaries which might vary in terms of length, or the types of aspects
being mentioned, rendering the application of transfer learning
techniques \cite{Pan:ea:2010} problematic.

Motivated by these issues, \citet{chu2019meansum} consider an
\emph{unsupervised} learning setting where there are only documents
(product or business reviews) available without corresponding
summaries. They propose an end-to-end neural model to perform
abstractive summarization based on (a)~an autoencoder that learns
representations for each review and (b)~a summarization module which
takes the aggregate encoding of reviews as input and learns to
generate a summary which is semantically similar to the source
documents.  Due to the absence of ground truth summaries, the model is
not trained to reconstruct the aggregate encoding of reviews, but
rather it only learns to reconstruct the encoding of \emph{individual}
reviews. As a result, it may not be able to generate meaningful text
when the number of reviews is large. Furthermore, autoencoders are
constrained to use simple decoders lacking attention
\cite{bahdanau2015neural} and copy \cite{vinyals2015pointer}
mechanisms which have proven useful in the supervised setting leading
to the generation of informative and detailed
summaries. Problematically, a powerful decoder might be detrimental to
the reconstruction objective, learning to express arbitrary distributions of
the output sequence while ignoring the encoded input
\cite{kingma2014auto,bowman-etal-2016-generating}.

%ruct the aggregate
%ns to reconstruct
% may not be able to
%reviews is large.
%e simple decoders, since
%015neural} and copy
%coders stronger and
%encoder
%\cite{kingma2014auto}. 

%A major obstacle for supervised approaches is that in most domains,
%training data in the form of review-summary pairs is not available.
%Moreover, manually annotating summaries is practically impossible
%since an annotator must read all available reviews to write an opinion
%summary, which can be prohibitively many. Thus, a supervised learning
%setup is not a realistic solution. In this paper, our goal is to solve
%abstractive opinion summarization in an unsupervised manner.
%, i.e., our model does not require human-annotated training data.

%\citet{chu2019meansum} recently proposed an unsupervised abstractive
%summarization model based on an autoencoder that generates an output
%summary using an aggregate encoding of input reviews. Their model is
%trained using two objectives. Firstly, a reconstruction loss on the
%reviews is used to learn how to generate coherent texts.  Secondly, a
%similarity loss of the generated summary and the aggregate encoding is
%used to learn how to generate texts that are semantically similar to
%the input reviews.  Their method suffers from two drawbacks, both are
%due to the absence of ground truth summaries.

%Firstly, training requires the use of sampling techniques \cite{jang2017categorical} to make the model fully differentiable. This slows optimization and induces bias due to function mismatch during backpropagation \cite{tucker2017rebar}.

In this paper, we enable the use of supervised techniques for
unsupervised summarization.  Specifically, we automatically generate a
synthetic training dataset from a corpus of product reviews, and use
this dataset to train a more powerful neural model with supervised
learning.  The synthetic data is created by selecting a review from
the corpus, pretending it is a summary, generating multiple noisy
versions thereof and treating these as \emph{pseudo-reviews}.  The
latter are obtained with two noise generation functions targeting
textual units of different granularity: \emph{segment} noising
introduces noise at the word- and phrase-level, while \emph{document}
noising replaces a review with a semantically similar one.  We use the
synthetic data to train a neural model that learns to denoise the
pseudo-reviews and generate the summary.  This is motivated by how
humans write opinion summaries, where denoising can be seen as
removing diverging information.  Our proposed model
consists of a multi-source encoder and a decoder equipped with an
attention mechanism.  Additionally, we introduce three modules:
(a)~explicit denoising guides how the model removes noise from the
input encodings, (b) partial copy enables to copy information from the
source reviews only when necessary, and (c) a discriminator helps the
decoder generate topically consistent text.

We perform experiments on two review datasets representing different
domains (movies vs businesses) and summarization requirements (short
vs longer summaries).  Results based on automatic and human evaluation
show that our method outperforms previous unsupervised summarization
models, including the state-of-the-art abstractive system of
\citet{chu2019meansum} and is on the same par with a state-of-the-art
supervised model \cite{wang2016neural} trained on a small sample of
(genuine) review-summary pairs.

%%%%%%%%%%%%%%%%%%%%%%%%%%%%%%%%%%%%%%%%%%%%%%%%%%%%%%%%%%%%%%%%%
\section{Related Work}
\label{sec:related-work}

%Recently, we have seen advances on summarization using supervised
%methods and neural models\footnote{We refer the readers to
%  \citet{nenkova2012survey} and \citet{lin2019abstractive} for surveys
%  of this line of topic.}. Most focus on single document summarization
%\cite{rush2015neural,nallapati2016abstractive}, with the exception of
%recent litearture on multi-document summarization of Wikipedia pages
%\cite{liu2018generating}, news articles \cite{fabbri2019multi} and
%movie reviews \cite{wang2016neural}.  In this work, we focus on
%unsupervised learning methods, which do not require training data.

%Summarizing opinions from reviews can either be structured or textual. Structured summaries include clustering reviews based on aspect-based sentiments \cite{hu2006opinion} and constructing a table showing the sentiment and terms used for each aspect \cite{lu2009rated}.
%Methods include pipeline systems of aspect extraction and sentiment analysis \cite{liu2005opinion,amplayo2017adaptable} and extensions to Dirichlet topic models \cite{titov2008joint,jo2011aspect}.

%Textual summaries are usually the desired output by humans since they are more compact and easier to digest \cite{mckeown2005do}. 

Most previous work on unsupervised opinion summarization has focused
on extractive approaches
\cite{carenini2006multi,ku2006opinion,paul2010summarizing,angelidis2018summarizing}
where a clustering model groups opinions of the same aspect, and a
sentence extraction model identifies text representative of each
cluster. \citet{ganesan2010opinosis} propose a graph-based abstractive
framework for generating concise opinion summaries, while
\citet{difabbrizio2014hybrid} use an extractive system to first select
salient sentences and then generate an abstractive summary based on
hand-written templates \cite{carenini2006generating}.

As mentioned earlier, we follow the setting of \citet{chu2019meansum}
in assuming that we have access to reviews but no gold-standard
summaries. Their model learns to generate opinion summaries by
reconstructing a canonical review of the average encoding of input
reviews. Our proposed method is also abstractive and neural-based, but
eschews the use of an autoencoder in favor of supervised
sequence-to-sequence learning through the creation of a synthetic
training dataset.  Concurrently with our work,
\citet{bravzinskas2019unsupervised} use a hierarchical variational
autoencoder to learn a latent code of the summary. While they also use
randomly sampled reviews for supervised training, our dataset
construction method is more principled making use of linguistically
motivated noise functions.

Our work relates to denoising autoencoders (DAEs;
\citealp{vincent2008extracting}), which have been effectively used as
unsupervised methods for various NLP tasks. Earlier approaches have
shown that DAEs can be used to learn high-level text representations
for domain adaptation \cite{glorot2011domain} and multimodal
representations of textual and visual input
\cite{silberer2014learning}.  Recent work has applied DAEs to text
generation tasks, specifically to data-to-text generation
\cite{freitag2018unsupervised} and extractive sentence compression
\cite{fevry2018unsupervised}. Our model differs from these approaches
in two respects. Firstly, while previous work has adopted trivial
noising methods such as randomly adding or removing words
\cite{fevry2018unsupervised} and randomly corrupting encodings
\cite{silberer2014learning}, our noise generators are more
linguistically informed and suitable for the opinion summarization
task. Secondly, while in \citet{freitag2018unsupervised} the decoder
is limited to vanilla RNNs, our noising method enables the use of more
complex architectures, enhanced with attention and copy mechanisms,
which are known to improve the performance of summarization systems
\cite{rush2015neural,see2017get}.

\section{Modeling Approach}
\label{sec:modeling-approach}

%In this paper, we focus on the task of unsupervised abstractive
%opinion summarization. 
Let $\mathbf{X} = \{x_1,...,x_N\}$ denote a set of reviews about a
product (e.g.,~a movie or business). Our aim is to generate a summary~$y$ of the
opinions expressed in~$\mathbf{X}$.  We further assume access to a
corpus $\mathbb{C} = \{\mathbf{X}_1,...,\mathbf{X}_M\}$ containing
multiple reviews about $M$ products without corresponding opinion
summaries. %, at training time.

%of multiple reviews about $M$ products $\mathbb{C} =
%\{\mathbf{X}_1,...,\mathbf{X}_M\}$, without corresponding opinion
%summaries, at training time.

Our method consists of two parts. We first create a synthetic
dataset~$\mathbb{D}=\{(\mathbf{X},y)\}$ consisting of summary-review
pairs. Specifically, we sample review~$x_i$ from $\mathbb{C}$, pretend
it is a summary, and generate multiple noisy versions thereof
(i.e.,~pseudo-reviews). At training time, a denoising model learns to
remove the noise from the reviews and generate the summary.  At test
time, the same denoising model is used to summarize actual reviews.
We use denoising as an auxiliary task for opinion summarization to
simulate the fact that summaries tend to omit opinions that do not
represent consensus (i.e.,~noise in the pseudo-review), but include
salient opinions found in most reviews (i.e.,~non-noisy parts of the
pseudo-review).  %We explain our method in detail below.

\subsection{Synthetic Dataset Creation via Noising}
\label{sec:noising}

\begin{figure}[t]
    \centering
    \includegraphics[width=\columnwidth]{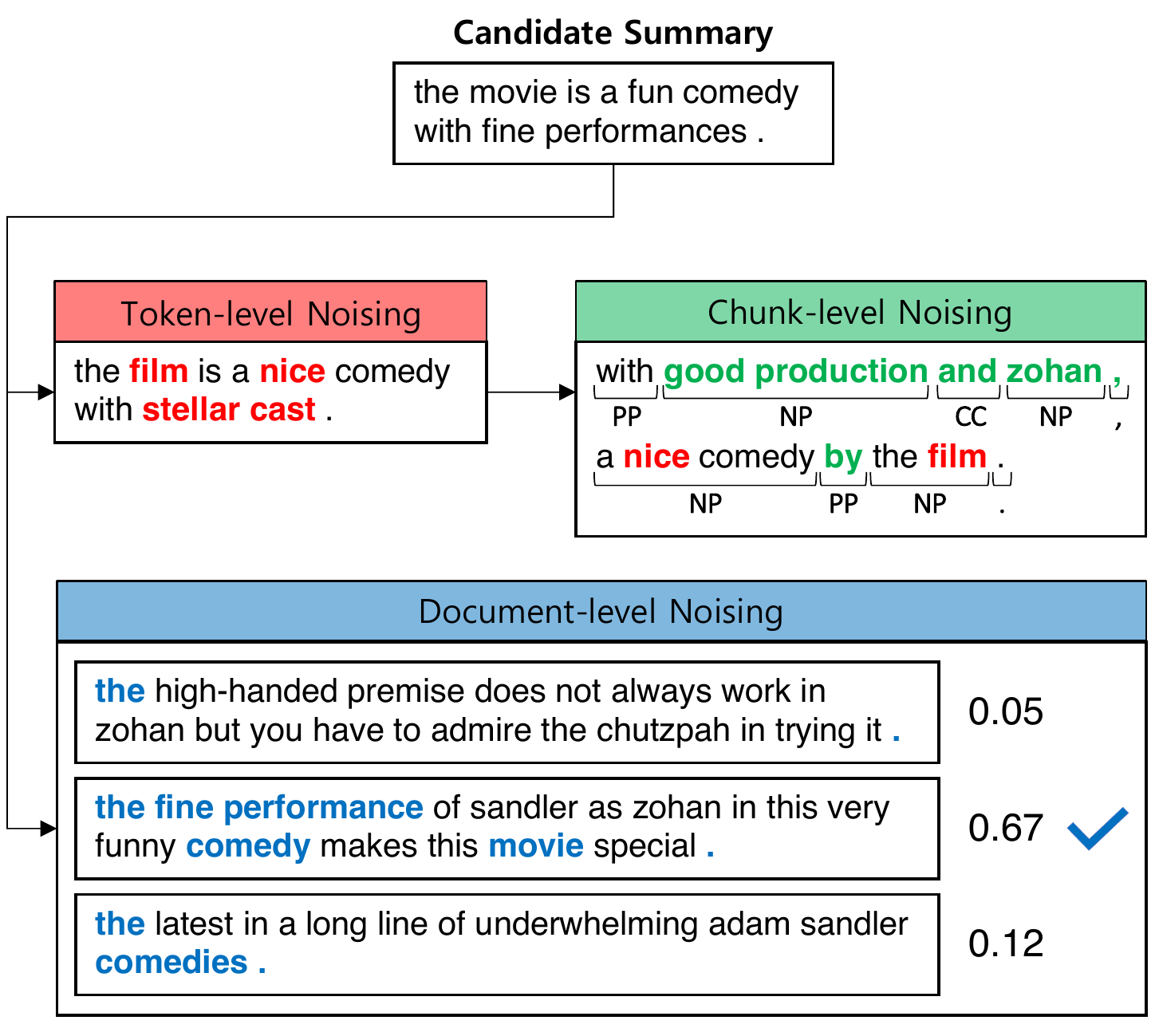}
    \caption{Synthetic dataset creation.  Given a sampled candidate
      summary, we add noise using two methods: (a) segment noising
      performs token- and chunk-level alterations, and (b) document
      noising replaces the text with a semantically similar review.}
    \label{fig:noising}
\end{figure}

We sample a review as a candidate summary and generate noisy versions
thereof, using two functions: (a) segment noising adds noise at the
token and chunk level, and (b) document noising adds noise at the text
level.  The noise functions are illustrated in
Figure~\ref{fig:noising}.

\paragraph{Summary Sampling}

Summaries and reviews follow different writing conventions. For
example, reviews are subjective, and often include first-person
singular pronouns such as \textit{I} and \textit{my} and several
unnecessary characters or symbols. They may also vary in length and
detail. We discard reviews from corpus $\mathbb{C}$ which display an
excess of these characteristics based on a list of domain-specific
constraints (detailed in Section \ref{sec:setup}). We sample a
review~$y$ from the filtered corpus, which we use as the candidate
summary.

%We thus filter
%out unsuitable reviews from the corpus $\mathbb{C}$ based on a
%domain-specific list of constraints.  We list these constraints, which
%differ for each dataset, in Section \ref{sec:setup}.  Finally, we
%sample a review $y$ from the filtered corpus, which we use as the
%candidate summary.

\paragraph{Segment Noising}

Given candidate summary \mbox{$y=\{w_1,...,w_L\}$}, we create a set of
segment-level noisy versions $\mathbf{X}^{(c)} = \{x^{(c)}_1,...,x^{(c)}_N\}$.
Previous work has adopted noising techniques based on random
\mbox{$n$-gram} alterations \cite{fevry2018unsupervised}, however, we
instead rely on two simple, linguistically informed noise functions.
Firstly, we train a bidirectional language model (BiLM;
\citealp{peters2018deep}) on the review corpus~$\mathbb{C}$.  For each
word in~$y$, the BiLM predicts a softmax word distribution which can be
used to replace words.  Secondly, we utilize
FLAIR\footnote{\url{https://github.com/zalandoresearch/flair}}
\cite{akbik2019flair}, an off-the-shelf state-of-the-art syntactic
chunker that leverages contextual embeddings, to shallow parse each
review $r$ in corpus $\mathbb{C}$.  This results in a list of chunks
$\mathbf{C}_r = \{c_1, ..., c_K\}$ with corresponding syntactic labels
$\mathbf{G}_r = \{g_1, ..., g_K\}$ for each review $r$, which we use
for replacing and rearranging chunks.

Segment-level noise involves token- and chunk-level alterations.
Token-level alterations are performed by replacing tokens in~$y$ with
probability~$p^{\mathcal{R}}$.  Specifically, we replace token~$w_j$
in~$y$, by sampling token~$w'_j$ from the BiLM predicted word
distribution (see in Figure~\ref{fig:noising}).  We use nucleus
sampling \cite{hotzman2019curious}, which samples from a rescaled
distribution of words with probability higher than a threshold
$p^{\mathcal{N}}$, instead of the original distribution.  This has
been shown to yield better samples in comparison to top-$k$ sampling,
mitigating the problem of text degeneration \cite{hotzman2019curious}.

Chunk-level alterations are performed by removing and inserting chunks
in $y$, and rearranging them based on a sampled syntactic template.
Specifically, we first shallow parse~$y$ using FLAIR, obtaining a list
of chunks $\mathbf{C}_y$, each of which is removed with
probability~$p^{\mathcal{R}}$. We then randomly sample a review~$r$
from our corpus and use its sequence of chunk labels~$\mathbf{G}_r$ as
a syntactic template, which we fill in with chunks in $\mathbf{C}_y$
(sampled without replacement), if available, or with chunks in corpus
$\mathbb{C}$, otherwise.  This results in a noisy version $x^{(c)}$
(see Figure~\ref{fig:noising} for an example). Repeating the process
$N$ times produces the noisy set $\mathbf{X}^{(c)}$. We describe this
process step-by-step in the Appendix.

%
%We then remove chunks~$c_k \in
%\mathbf{C}_y$, we
%remove it with a probability $p_{remove}$, i.e., $\mathbf{C}_y =
%\mathbf{C}_y - \{c_k\}$.  Then, we randomly sample a review $r$ in the
%corpus and use its list of chunk tags $\mathbf{G}_r$ as the syntactic
%template.  Finally, for each chunk tag $g_k \in \mathbf{G}_r$, we
%sample without replacement from $\mathbf{C}_y$ if there exists a chunk
%with tag $g_k$ in $y$, otherwise we sample from the whole review
%corpus.  The new list of chunks is concatenated together into one text
%sequence, resulting to a noisy version $x^{(c)}$. Repeating the
%process $N$ times produces the noisy set $\mathbf{X}^{(c)}$

\paragraph{Document Noising}

Given candidate summary $y=\{w_1,...,w_L\}$, we also create another
set of document-level noisy versions $\mathbf{X}^{(d)} =
\{x^{(d)}_1,...,x^{(d)}_N\}$.  Instead of manipulating parts of the
summary, we altogether replace it with a similar review from the
corpus and treat it as a noisy version.  Specifically, we select
$N$~reviews that are most similar to~$y$ and discuss the same product.
To measure similarity, we use IDF-weighted \mbox{ROUGE-1} F1
\cite{lin2004rouge}, where we calculate the lexical overlap between
the review and the candidate summary, weighted by token importance:
\begin{gather*}
    {overlap} = \sum_{w_j \in x} \big( \text{IDF}(w_j) * 1(w_j \in y) \big)\\
    \text{P} = {overlap}/{|x|}\hspace*{1cm} \quad \text{R} = {overlap}/{|y|} \\
    \text{F}_{1} = {(2*\text{P}*\text{R})}/{(\text{P}+\text{R})}
\end{gather*}
where $x$ is a review in the corpus, $1(\cdot)$ is an indicator
function, and P, R, and F$_1$ are the \mbox{ROUGE-1} precision,
recall, and F$_{1}$, respectively.  The reviews with the highest
F$_{1}$ are selected as noisy versions of~$y$, resulting in the noisy
set $\mathbf{X}^{(d)}$ (see Figure~\ref{fig:noising}).

We create a total of~$2*N$ noisy versions of~$y$,
i.e.,~$\mathbf{X} = \mathbf{X}^{(c)} \cup \mathbf{X}^{(d)}$ and obtain
our synthetic training data $\mathbb{D} = \{(\mathbf{X}, y)\}$ by
generating $|\mathbb{D}|$ pseudo-review-summary pairs.  Both noising
methods are necessary to achieve aspect diversity amongst input
reviews. Segment noising creates reviews which may mention aspects not
found in the summary, while document noising creates reviews with
content similar to the summary. Relying on either noise function alone
decreases performance (see the ablation studies in
Section~\ref{sec:exp}).  We show examples of these noisy versions in
the Appendix.

\subsection{Summarization via Denoising}

\begin{figure*}[t]
    \centering
    \includegraphics[width=\textwidth]{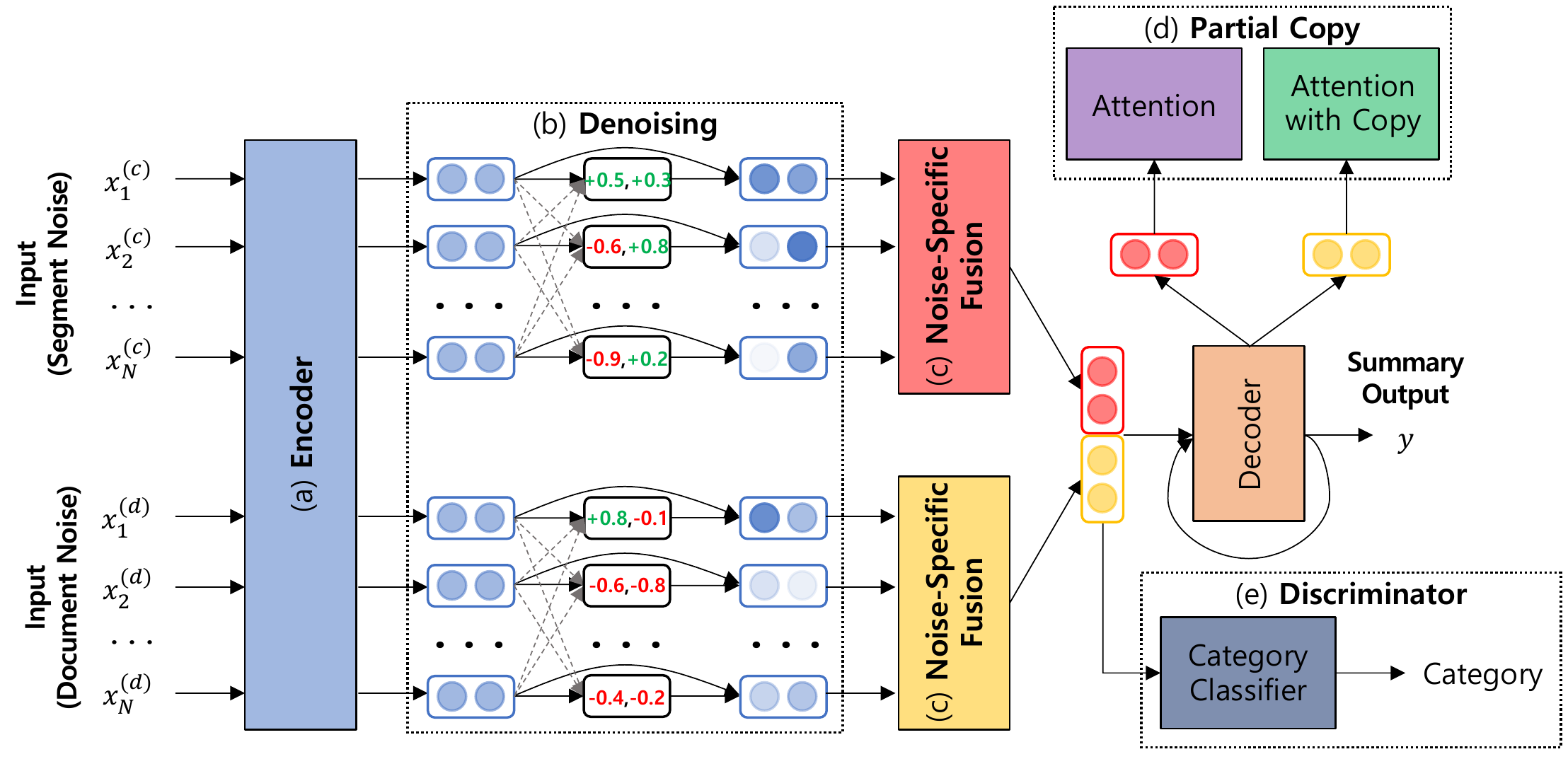}
    \caption{Architecture of \textsc{DenoiseSum}: it consists of a
      multi-source encoder with explicit denoising, noise-specific
      fusion, a decoder with partial copy, and a review category
      classifier.}
    \label{fig:denoising}
\end{figure*}

We summarize (aka denoise) the input~$\mathbf{X}$ with our model which
we call \textsc{DenoiseSum}, illustrated in Figure~\ref{fig:denoising}. 
A multi-source encoder produces an
encoding for each pseudo-review.  The encodings are further corrected
via an explicit denoising module, and then fused into an aggregate
encoding for each type of noise.  Finally, the fused encodings are
passed to a decoder with a partial copy mechanism to generate the
summary~$y$.

\paragraph{Multi-Source Encoder}

For each pseudo-review \mbox{$x_j \in \mathbf{X}$} where
$x_j=\{w_1,...,w_L\}$ and $w_k$ is the $k$th token in $x_j$, we obtain
contextualized token encodings~$\{h_k\}$ and an overall review
encoding~$d_j$ with a BiLSTM encoder \cite{hochreiter1997long}:
\begin{align*}
    \overrightarrow{h}_k &= \text{LSTM}_f(w_k,\overrightarrow{h}_{k-1}) \\
    \overleftarrow{h}_k &= \text{LSTM}_b(w_k,\overleftarrow{h}_{k+1}) \\
    h_k &= [\overrightarrow{h}_k;\overleftarrow{h}_k] \\
    d_j &= [\overrightarrow{h}_L;\overleftarrow{h}_1]
\end{align*}
where $\overrightarrow{h}_k$ and $\overleftarrow{h}_k$ are forward and
backward hidden states of the BiLSTM at timestep $k$, and~$;$ denotes
concatenation (see module~(a) in Figure~\ref{fig:denoising}). 

\paragraph{Explicit Denoising}

The model should be able to remove noise from the encodings before
decoding the text. While previous methods
\cite{vincent2008extracting,freitag2018unsupervised} \emph{implicitly}
assign the denoising task to the encoder, we propose an
\emph{explicit} denoising component (see module~(b) in
Figure~\ref{fig:denoising}).  Specifically, we create a correction
vector~$c^{(c)}_j$ for each pseudo-review~$d^{(c)}_j$ which resulted
from the application of segment noise.  $c^{(c)}_j$~represents the
adjustment needed to denoise each dimension of ~$d^{(c)}_j$ and is
used to create~$\hat{d}^{(c)}_j$, a denoised encoding of~$d^{(c)}_j$:
\begin{align*}
    q &= \sum_{j=1}^N d^{(c)}_j / N \\
    c^{(c)}_j &= \tanh (W^{(c)}_d [d^{(c)}_j; q] + b^{(c)}_d) \\
    \hat{d}^{(c)}_j &= d^{(c)}_j + c^{(c)}_j
\end{align*}
where $q$~represents a mean review encoding and functions as a query
vector, $W$~and~$b$ are learned parameters, and superscript~${}^{(c)}$
signifies segment noising. We can interpret the correction
vector as removing or adding information to each dimension when its
value is negative or positive, respectively.  Analogously, we
obtain~$\hat{d}^{(d)}_j$ for pseudo-reviews~$d^{(d)}_j$ which have
been created with document noising. 
% to obtain $\hat{d}^{(d)}_j$.%, where ${}^(d)$ signifies
%document-level noising.
%We also tried using importance gates \cite{} to correct the vectors, however we found our method to have better performance.

\paragraph{Noise-Specific Fusion}

For each type of noise (segment and document), we create a
noise-specific aggregate encoding by fusing the denoised encodings
into one (see module~(c) in Figure~\ref{fig:denoising}).
Given~$\{\hat{d}^{(c)}_j\}$, the set of denoised encodings
corresponding to segment noisy inputs, we create aggregate
encoding~$s^{(c)}_0$:
\begin{align*}
    \alpha^{(c)}_j &= \text{softmax} (W^{(c)}_f \hat{d}^{(c)}_j + b^{(c)}_f) \\
    s^{(c)}_0 &= \sum_j \hat{d}^{(c)}_j * \alpha^{(c)}_j
\end{align*}
where~$\alpha_j$ is a gate vector with the same dimensionality as the
denoised encodings. Analogously, we obtain~$s^{(d)}_0$ from the
denoised encodings~$\{\hat{d}^{(d)}_j\}$ corresponding to
document noisy inputs.

\paragraph{Decoder with Partial Copy}
Our decoder generates a summary given encodings~$s^{(c)}_0$
and~$s^{(d)}_0$ as input. An advantage of our method is its ability to
incorporate techniques used in supervised models, such as attention
\cite{bahdanau2015neural} and copy \cite{vinyals2015pointer}.
Pseudo-reviews created using segment noising include various chunk
permutations, which could result to ungrammatical and incoherent text.
Using a copy mechanism on these texts may hurt the fluency of the
output.  We therefore allow copy on document noisy
inputs only (see module~(d) in Figure~\ref{fig:denoising}).

We use two LSTM decoders for the aggregate encodings, one equipped
with attention and copy mechanisms, and one without copy mechanism.
We then combine the results of these decoders using a learned gate.
Specifically, token~$w_t$ at timestep~$t$ is predicted as:
\begin{align*}
    s^{(c)}_t, p^{(c)}(w_t) &= \text{LSTM}_\text{att} (w_{t-1}, s^{(c)}_{t-1}) \\
    s^{(d)}_t, p^{(d)}(w_t) &= \text{LSTM}_\text{att+copy} (w_{t-1}, s^{(d)}_{t-1}) \\
    \lambda_t &= \sigma( W_p [w_{t-1}; s^{(c)}_t; s^{(d)}_t] + b_p) \\
    p(w_t) &= \lambda_t \hspace*{-.4ex}*\hspace*{-.4ex}p^{(c)}(w_t)  + (1-\lambda_t)\hspace*{-.4ex}*\hspace*{-.4ex}p^{(d)}(w_t)
\end{align*}
where $s_t$ and $p(w_t)$ are the hidden state and predicted token
distribution at timestep $t$, and $\sigma(\cdot)$ is the sigmoid
function.

\subsection{Training and Inference}

We use a maximum likelihood loss to optimize the generation
probability distribution based on summary $y = \{w_1, ...,
w_L\}$ from our synthetic dataset:
\begin{equation}
    \mathcal{L}_{gen} = - \sum_{w_t\in y} \log p(w_t) \nonumber
\end{equation}

The decoder depends on~$\mathcal{L}_{gen}$ to generate meaningful,
denoised outputs. As this is a rather \emph{indirect} way to optimize
our denoising module, we additionally use a discriminative loss
providing \emph{direct} supervision. The discriminator operates at the
output of the fusion module and predicts the category
distribution~$p(z)$ of the output summary~$y$ (see module (e) in
Figure \ref{fig:denoising}). The type of categories varies across
domains.  For movies, categories can be information about their genre
(e.g.,~drama, comedy), while for businesses their specific type
(e.g.,~restaurant, beauty parlor). This information is often included
in reviews but we assume otherwise and use an LDA topic model
\cite{blei2003latent} to infer $p(z)$
%can also be inferred when unavailable, e.g.,~with an LDA
%topic model \cite{blei2003latent} 
(we present experiments with human labeled and automatically induced
categories in Section~\ref{sec:exp}). An MLP classifier takes as input
aggregate encodings~$s^{(c)}$ and~$s^{(d)}$ and infers~$q(z)$. The
discriminator is trained by calculating the KL divergence between
predicted and actual category distributions~$q(z)$ and~$p(z)$:
% where we append an 
% Specifically,
%assume we know the category distribution $p(z)$ of the candidate
%summary~$y$.  This  be either a human-annotated category, if
%available, such as the movie genre or the product category, or a topic
%distribution induced by an LDA topic model \cite{blei2003latent}.  Our
%experiments in Section \ref{sec:exp} show that the latter is a better
%supervision, which means that reliance towards human categories is 
%unnecessary.  We append an MLP classifier that learns to infer the
%category distribution through $q(z)$, given the aggregated encodings
%$s^{(c)}$ and $s^{(d)}$. Finally, we calculate the KL divergence
%between $p(z)$ and $q(z)$:
%
\begin{align*}
    q(z) &= \text{MLP}_d (s^{(c)}, s^{(d)}) \\
    \mathcal{L}_{disc} &= \infdiv{p(z)}{q(z)}
\end{align*}
The final objective is the sum of both loss functions:
\begin{equation}
\mathcal{L} = \mathcal{L}_{gen} + \mathcal{L}_{disc} \nonumber
\end{equation}

At test time, we are given genuine reviews~$\mathbf{X}$ as input
instead of the synthetic ones. We generate a summary by 
%pretending
treating
$\mathbf{X}$ 
%is noisy and treating it 
as~$\mathbf{X}^{(c)}$
and~$\mathbf{X}^{(d)}$, i.e.,~the outcome of segment and
document noising.
%noise~$\mathbf{X}^{(c)}$ and document-level noise~$\mathbf{X}^{(d)}$.
%When the number of
%input reviews $|\mathbf{X}| > N$, instead of using an extractive stage
%that selects the top $N$ reviews, we find using all reviews as input
%to have better performance.

%%%%%%%%%%%%%%%%%%%%%%%%%%%%%%%%%%%%%%%%%%%%%%%%%%%%%%%%%%%%%%%%%%%%%%%%%%
\section{Experimental Setup}
\label{sec:setup}

\begin{table}[t]
%    \small
    \centering
    \begin{tabular}{@{~~}lrrr@{~~}}\thickhline
      Rotten Tomatoes    &    Train* & Dev &  Test \\      \thickhline
        \#movies & 25k  & 536 & 737 \\
        \#reviews/movie &  40.0 & 98.0 & 100.3 \\
        \#tokens/review & 28.4 & 23.5 & 23.6 \\
        \#tokens/summary & 22.7 & 23.6 & 23.8 \\
   corpus size & \multicolumn{3}{r}{245,848} \\
        \thickhline
        \multicolumn{4}{c}{} \\
        \thickhline
      Yelp  & Train* & Dev & Test \\
        \thickhline
        \#businesses & 100k & 100 & 100 \\
        \#reviews/business & 8.0 & 8.0 & 8.0 \\
        \#tokens/review & 72.3 & 70.3 & 67.8 \\
        \#tokens/summary & 64.8 & 70.9 & 67.3 \\
        corpus size & \multicolumn{3}{r}{2,320,800} \\
        \thickhline
    \end{tabular}
    \caption{Dataset statistics; Train* column refers to the synthetic
      data we created through noising (Section \ref{sec:noising}).} 
    \label{tab:datasets}
\end{table}

\begin{table*}[t]
%  \small
  \centering
\begin{tabular}{@{}l@{\hspace{.6cm}}l@{}}
\begin{minipage}[]{3.5in}
\centering{
    \begin{tabular}{@{}l@{~~}c@{~~}c@{~~}c@{~~}c@{~~}c@{}}
    \thickhline
    \multicolumn{1}{c}{Model} & METEOR & RSU4  & R1    & R2    & RL \\
    \thickhline
\textsc{Oracle} & \hspace*{-1ex}12.10 & \hspace*{-1ex}12.01 & 30.94 & 10.75 & 24.95 \\
   \textsc{LexRank}* & 5.59  & 3.98  & ---     & ---     & --- \\
   \textsc{Word2Vec} & 6.14 & 4.04  & 13.93     & 2.10     & 10.81 \\
   \textsc{SentiNeuron} & 7.02  & 4.77  & 15.90     & 2.01     & 11.74 \\
    \hline
    \textsc{Opinosis}* & 6.07  & 4.90  & ---     & ---     & --- \\
    \textsc{MeanSum} & 6.07  & 4.41  & 15.79     & 1.94     & 12.26 \\
    \textsc{DenoiseSum} & \textbf{8.30}  & \textbf{6.84}  & \textbf{21.26}     & \textbf{4.61}     & \textbf{16.27} \\
    \thickhline
%    \rowcolor{Gray}
    Best Supervised* & 8.50  & 7.39  & 21.19     & 7.64     & 17.80 \\
    \thickhline
    \end{tabular}%
    \caption{Automatic evaluation on \textbf{Rotten Tomatoes}.
      Results from \citet{amplayo2019informative} are marked with an
      asterisk~*. Extractive/abstractive models shown in the
      first/second block. Best performing results for unsupervised
      models are \textbf{boldfaced}.}
  \label{tab:rotten}%
}
\end{minipage} & 
\begin{minipage}[]{2.6in}
\centering{
    \begin{tabular}{@{~~}lccc@{~~}}
    \thickhline
    \multicolumn{1}{c}{Model} &  R1    & R2    & RL \\
    \thickhline
\textsc{Oracle} & 31.07 & 6.11 & 18.11 \\
   \textsc{LexRank} & 24.62 & 3.66 & 14.51 \\
   \textsc{Word2Vec}* & 24.61 & 2.85  & 13.81    \\
   \textsc{SentiNeuron} & 25.05  & 3.09  & 14.56   \\
    \hline
    \textsc{Opinosis} & 20.85 & 1.52 & 11.46 \\
    \textsc{MeanSum}* & 28.86  & 3.66  & 15.91    \\
    \textsc{DenoiseSum} & \textbf{30.14}  & \textbf{4.99}  & \textbf{17.65} \\
    \thickhline
    \end{tabular}%
    \caption{Automatic evaluation on \textbf{Yelp}.  Results 
      from \citet{chu2019meansum} are marked with an
      asterisk~*. Extractive/abstractive models shown in the
      first/second block.  Best performing unsupervised models are
      \textbf{boldfaced}.}
  \label{tab:yelp}%
}
\end{minipage}
\end{tabular}
\end{table*}%

\paragraph{Dataset}

We performed experiments on two datasets which represent different
domains and summary types.  The Rotten Tomatoes
dataset\footnote{\url{http://www.ccs.neu.edu/home/luwang/data.html}}
\cite{wang2016neural} contains a large set of reviews for various
movies written by critics. Each set of reviews has a
gold-standard consensus summary written by an editor. We follow the
partition of \citet{wang2016neural} but do not use ground truth
summaries during training to simulate our unsupervised setting.  The
Yelp dataset\footnote{\url{https://github.com/sosuperic/MeanSum}} in
\citet{chu2019meansum} includes a large training corpus of reviews
without gold-standard summaries. The latter are provided for the
development and test set and were generated by an Amazon Mechanical
Turker. We follow the splits introduced in their work. A comparison
between the two datasets is provided in Table~\ref{tab:datasets}. As
can be seen, Rotten Tomatoes summaries are generally short,
%(approximately 23 tokens),
while Yelp reviews are three times longer.
%(between 70 and 67 tokens). 
Interestingly, there are a lot more reviews to summarize in Rotten
Tomatoes (approximately 100~reviews) while input reviews in Yelp are
considerably less (i.e.,~8 reviews).

%The Rotten Tomatoes dataset consists of
%movies, each of which contains a large set of reviews and a summary
%professionally written by an editor.  It originally contains a
%separate training split, however we only use the reviews as a single
%large corpus and disregard ground truth summaries to simulate an
%unsupervised setting.  Following previous work
%\cite{wang2016neural,amplayo2019informative}, we used a generic label
%for movie titles during training, which we replace with the original
%titles after generating the summary.  The Yelp dataset includes a
%large corpus of reviews without summaries, and development and test
%splits, where the summary is generated by an Amazon turker. Statistics
%are shown in Table \ref{tab:datasets}.

\paragraph{Implementation}

To create the synthetic dataset, we sample candidate summaries using
the following constraints: (1)~the number of non-alphanumeric symbols
must be less than~3, (2) there must be no first-person singular
pronouns (not used for Yelp), and (3)~the number of tokens must be
between 20 to~30 (50 to 90 for Yelp).  We set $p^{\mathcal{R}}$ to~0.8
and~0.4 for token and chunk noise, and
%for token noise and 0.4~for chunk noise, and
$p^{\mathcal{N}}$ to~0.9.  
For each review-summary pair, the number of reviews $N$ is sampled from the Gaussian distribution $\mathcal{N}(\mu, \sigma^2)$ where $\mu$ and $\sigma$ are the mean and standard deviation of the number of reviews in the development set.
We created 25k (Rotten Tomatoes) and 100k
(Yelp) pseudo-reviews for our synthetic datasets (see
Table~\ref{tab:datasets}).

We set the dimensions of the word embeddings to 300, the vocabulary
size to 50k, the hidden dimensions to 256, the batch size to 8, and
dropout \cite{srivastava2014dropout} to 0.1. For our discriminator, 
%we used the movie genre and business type categories available in the
%original datasets. We also 
we employed an LDA topic model trained on the review corpus,
%to infer $p(z)$, 
with 50 (Rotten Tomatoes) and 100 (Yelp) topics (tuned on the
development set).  The LSTM weights were pretrained with a language
modeling objective, using the corpus as training data.  For Yelp, we
additionally trained a coverage mechanism \cite{see2017get} in a
separate training phase to avoid repetition. We used the Adam
optimizer \cite{kingma2015adam} with a learning rate of 0.001 and
$l_2$ constraint of 3.  At test time, summaries were generated using
length normalized beam search with a beam size of~5. We performed
early stopping based on the performance of the model on the
development set. Our model was trained on a single GeForce GTX 1080 Ti
GPU and is implemented using PyTorch.\footnote{Our code can be
  downloaded from \url{https://github.com/rktamplayo/DenoiseSum}.}

\paragraph{Comparison Systems}

We compared \textsc{DenoiseSum} to several unsupervised extractive and
abstractive methods. Extractive approaches include
(a)~\textsc{LexRank} \cite{erkan2004lexrank}, an algorithm similar to
PageRank that generates summaries by selecting the most salient
sentences, (b)~\textsc{Word2Vec} \cite{rossiello2017centroid}, a
centroid-based method which represents the input as IDF-weighted word
embeddings and selects as summary the review closest to the centroid,
and (c)~\textsc{SentiNeuron}, which is similar to \textsc{Word2vec}
but uses a language model called Sentiment Neuron
\cite{radford2017learning} as input representation. As an upper bound,
\textsc{Oracle} selects as summary the review which maximizes the
ROUGE-1/2/L F1 score against the gold summary.

Abstractive methods include (d) \textsc{Opinosis}
\cite{ganesan2010opinosis}, a graph-based summarizer that generates
concise summaries of highly redundant opinions, and (e)
\textsc{MeanSum} \cite{chu2019meansum}, a neural
model that generates a summary by reconstructing
text from aggregate encodings of reviews. Finally, for Rotten
Tomatoes, we also compared with the state-of-the-art supervised model
proposed in \citet{amplayo2019informative} which used the original
training split.  Examples of system summaries are shown
in the Appendix.

%%%%%%%%%%%%%%%%%%%%%%%%%%%%%%%%%%%%%%%%%%%%%%%%%%%%%%%%%%%%%%%%%
\section{Results}
\label{sec:exp}

\begin{table}[t]
%  \small
  \centering
    \begin{tabular}{@{~~}lcc@{~~}}
    \thickhline
    \multicolumn{1}{c}{Model} & RT   & Yelp \\
    \thickhline
    \textsc{DenoiseSum} & 16.27 & 17.65 \\
    \hline
    \quad 10\% synthetic dataset & 15.39 & 16.22 \\
    \quad 50\% synthetic dataset & 15.76 & 17.54 \\
    \hline
    \quad no segment noising & 16.03 & 16.88 \\
    \quad no document noising & 16.22 & 16.67 \\
    \hline
    \quad no explicit denoising & 16.06 & 17.06 \\
    \quad no partial copy & 15.89 & 16.31 \\
    \quad no discriminator & 15.84 & 16.64 \\
    \quad using human categories & 15.87 & 15.86 \\
    \thickhline
    \end{tabular}%
    \caption{ROUGE-L of our model and versions thereof with less synthetic data (second block), using only one noising method (third block), and without some modules (fourth block). A more comprehensive table and discussion can be found in the Appendix.}
  \label{tab:ablation}%
\end{table}%

\begin{table*}[t]
%  \small
  \centering
    \begin{tabular}{@{}l@{~~}r@{~~}r@{~~}r@{~~}r@{~~}r@{~~}r@{~~}r@{}}
      \thickhline
      & \multicolumn{3}{c}{RT}   & \multicolumn{3}{c}{Yelp} \\
      \multicolumn{1}{c}{Model} & Inf & Coh & Gram & Inf & Coh & Gram \\
      \thickhline
      \textsc{SentiNeuron} & 11.8 & 8.3 & \textbf{25.4} & -24.8 & -0.8 & \textbf{9.3} \\
      \textsc{MeanSum} & -32.1 & -34.4 & -46.8 & 6.3 & -7.5 & -10.8 \\
      \textsc{DenoiseSum} & \textbf{20.3} & \textbf{26.1} &21.4 & \textbf{18.5} & \textbf{8.2} & 1.6\\
      \thickhline
    \end{tabular}%
   {\hspace{0.3cm}}
    \begin{tabular}{@{}l@{~~}c@{~~}c@{~~}c@{}}
    \thickhline
      \multicolumn{4}{c}{Yelp} \\
    \multicolumn{1}{c}{Model} & FullSupp & PartSupp & NoSupp\\
    \thickhline
    \textsc{MeanSum} & 41.7\% & 20.4\% & 38.0\% \\
    \textsc{DenoiseSum} & 55.1\% & 24.3\% & 20.5\% \\
    \textsc{Gold} & 63.6\% & 23.6\% & 12.8\% \\
    \thickhline
    \end{tabular}%
    \caption{Best-worst scaling (left) and summary veridicality
      (right) evaluation. Between systems differences are all significant,
      using a one-way ANOVA
      with posthoc Tukey HSD tests ($p<0.01$).} 
  \label{tab:human}%
\end{table*}%

\paragraph{Automatic Evaluation}

Our results on Rotten Tomatoes are shown in Table~\ref{tab:rotten}.
Following previous work \cite{wang2016neural,amplayo2019informative}
we report five metrics: METEOR \cite{denkowski2014meteor}, a
recall-oriented metric that rewards matching stems, synonyms, and
paraphrases; ROUGE-SU4 \cite{lin2004rouge}, the recall of unigrams and
skip-bigrams of up to four words; and the F1-score of ROUGE-1/2/L,
which respectively measures word-overlap, bigram-overlap, and the
longest common subsequence between system and reference summaries.
Results on Yelp are given in Table~\ref{tab:yelp} where we compare
systems using ROUGE-1/2/L F1, following \citet{chu2019meansum}.

As can be seen, \textsc{DenoiseSum} outperforms all competing models
on both datasets.  When compared to \textsc{MeanSum}, the difference
in performance is especially large on Rotten Tomatoes, where we see a
4.01 improvement in ROUGE-L. We believe this is because
\textsc{MeanSum} does not learn to reconstruct encodings of aggregated
inputs, and as a result it is unable to produce meaningful summaries
when the number of input reviews is large, as is the case for Rotten
Tomatoes.  In fact, the best extractive model, \textsc{SentiNeuron},
slightly outperforms \textsc{MeanSum} on this dataset across metrics
with the exception of ROUGE-L.  When compared to the best supervised
system, \textsc{DenoiseSum} performs comparably on several metrics,
specifically METEOR and ROUGE-1, however there is still a gap on
ROUGE-2, showing the limitations of systems trained without
gold-standard summaries.

%For Yelp, we compare systems
%using ROUGE-1/2/L F1-score, following \citet{chu2019meansum}.

Table~\ref{tab:ablation} presents various ablation studies on Rotten
Tomatoes (RT) and Yelp which assess the contribution of different
model components. Our experiments confirm that increasing the size of
the synthetic data improves performance, and that both segment and
document noising are useful. We also show that explicit denoising,
partial copy, and the discriminator help achieve best
results. Finally, human-labeled categories (instead of LDA topics)
decrease model performance, which suggests that more useful labels can
be approximated by automatic means.

\paragraph{Human Evaluation}

%In addition to automatic evaluation, 
We also conducted two judgment
elicitation studies using the Amazon Mechanical Turk (AMT)
crowdsourcing platform.  The first study assessed the quality of the
summaries using Best-Worst Scaling (BWS; \citealp{louviere2015best}),
a less labor-intensive alternative to paired comparisons that has been
shown to produce more reliable results than rating scales
\cite{kiritchenko2017best}.  Specifically, participants were shown the
movie/business name, some basic background information, and a
gold-standard summary.  They were also presented with three system
summaries, produced by \textsc{SentiNeuron} (best extractive model),
\textsc{MeanSum} (most related unsupervised model), and
\textsc{DenoiseSum}. 

Participants were asked to select the \textit{best} and \textit{worst}
among system summaries taking into account how much they deviated from
the ground truth summary in terms of: \emph{Informativeness}
(i.e.,~does the summary present opinions about specific aspects of the
movie/business in a concise manner?), \emph{Coherence} (i.e., is the
summary easy to read and does it follow a natural ordering of facts?),
and \emph{Grammaticality} (i.e.,~is the summary fluent and
grammatical?).  We randomly selected 50 instances from the test set.
We collected five judgments for each comparison. The order
of summaries was randomized per participant.  A rating per
system was computed as the percentage of times it was chosen as best
minus the percentage of times it was selected as worst. Results are
reported in Table~\ref{tab:human}, where Inf, Coh, and Gram are
shorthands for Informativeness, Coherence, and Grammaticality.
\textsc{DenoiseSum} was ranked best in terms of informativeness and
coherence, while the extractive system \textsc{SentiNeuron} was ranked
best on grammaticality. This is not entirely surprising since
extractive summaries written by humans are by definition grammatical.

Our second study examined the veridicality of the generated summaries,
namely whether the facts mentioned in them are indeed discussed in the
input reviews.  Participants were shown reviews and the corresponding
summary and were asked to verify for each summary sentence whether it
was fully supported by the reviews, partially supported, or not at all
supported.  We performed this experiment on Yelp only since the number
of reviews is small and participants could read them all in a timely
fashion. We used the same 50~instances as in our first study and
collected five judgments per instance.  Participants assessed the
summaries produced by \textsc{MeanSum} and \textsc{DenoiseSum}. We
also included \textsc{Gold}-standard summaries as an upper bound but
no output from an extractive system as it by default contains facts
mentioned in the reviews.

Table~\ref{tab:human} reports the percentage of fully (FullSupp),
partially (PartSupp), and un-supported (NoSupp) sentences.  Gold
summaries display the highest percentage of fully supported sentences
(63.3\%), followed by \textsc{DenoiseSum} (55.1\%), and
\textsc{MeanSum} (41.7\%). These results are encouraging, indicating
that our model hallucinates to a lesser extent compared to
\textsc{MeanSum}.

\section{Conclusions}
\label{sec:conclusions}

We consider an unsupervised learning setting for opinion
summarization where there are only reviews
available without corresponding summaries. Our key insight is to
enable the use of supervised techniques by creating synthetic
review-summary pairs using noise generation methods. Our
summarization model, \textsc{DenoiseSum}, introduces explicit
denoising, partial copy, and discrimination modules which improve
overall summary quality, outperforming competitive systems by a wide
margin. In the future, we would like to model aspects and sentiment
more explicitly as well as apply some of the techniques presented here
to unsupervised single-document summarization.

\section*{Acknowledgments}

We thank the anonymous reviewers for their feedback.  We gratefully
acknowledge the support of the European Research Council
(Lapata, award number 681760). The first author is supported by a
Google PhD Fellowship.

\bibliography{acl2020}
\bibliographystyle{acl_natbib}

\clearpage
\appendix
\section{Appendix}

\subsection{Segment Noising}

Algorithm~\ref{alg:seg_noising} shows how segment noising
(i.e.,~token- and chunk-level alterations) is applied step-by-step
(see Section~\ref{sec:noising}).  Segment noising assumes we have
access to a language model~$LM$ that is able to output token-level
predictions given neighboring tokens, and a syntactic chunker~$SC$
that is able to return shallow parses (i.e., chunks with corresponding
syntactic labels).  Function \textsc{TokenAlter} takes candidate
summary~$y$ as input and generates token-level alterations. The noisy
summary is then passed on to function \textsc{ChunkAlter} to create
chunk-level alterations. Sequentially executing both functions
produces noisy version~$\mathbf{X}^{(c)}$.

\alglanguage{pseudocode}
\begin{algorithm*}[h]
\small
\caption{Segment Noising}
\label{alg:seg_noising}

\begin{algorithmic}[1]
\Function{TokenAlter}{candidate summary $y$, language model $LM$, replace probability $p^\mathcal{R}$, nucleus threshold $p^\mathcal{N}$}
    \State Set noisy version $y' \gets \{\}$
    \For {$w_j \in y$}
        \State Set predicted word distribution $p(w) \gets LM(w_j)$
        \State Remove $w$ in $p(w)$ if $p(w) < p^\mathcal{N}$
        \State Set nucleus distribution $p^\mathcal{N}(w) \gets \frac{p(w)}{\sum_z p(z)}$
        \State Sample new token $w'_j$ from $p^\mathcal{N}(w)$
        \State Random sample a number $\hat{p}^\mathcal{R}$ between 0 and 1
        \If {$\hat{p}^\mathcal{R} < p^\mathcal{R}$} \Comment{\textit{replace token}}
        \State Update $y' \gets y' + \{w'_j\}$
        \Else
        \State Update $y' \gets y' + \{w_j\}$
        \EndIf 
    \EndFor
    \State \Return token-level noisy version $y'$
\EndFunction
\State {}
\Function{ChunkAlter}{candidate summary $y$, review corpus $\mathbb{C}$, syntactic chunker $SC$, remove probability $p^\mathcal{R}$}
    \State Set summary chunks and tags $\mathbf{C}_y, \mathbf{G}_y \gets SC(y)$
    \For {$c_y \in \mathbf{C}_y$}
        \State Random sample a number $\hat{p}^\mathcal{R}$ between 0 and 1
        \If {$\hat{p}^\mathcal{R} < p^\mathcal{R}$} \Comment{\textit{remove chunk}}
        \State Update $\mathbf{C}_y \gets \mathbf{C}_y - \{c_y\}$
        \EndIf 
    \EndFor
    \State Sample review $r$ from $\mathbb{C}$
    \State Set review chunks and tags $\mathbf{C}_r, \mathbf{G}_r \gets SC(r)$
    \State Set noisy version $y' \gets \{\}$
    \For {$g_r \in \mathbf{G}_r$}
        \If {$g_r \in \mathbf{G}_y$} \Comment{\textit{add chunk from original summary}}
            \State Sample (w/o replacement) chunk $c_y$ from $\mathbf{C}_y$ such that tag of $c_y$ is $g_r$
        \Else \Comment{\textit{add random chunk from corpus}}
            \State Sample chunk $c_y$ from corpus $\mathbb{C}$ such that tag of $c_y$ is $g_r$
        \EndIf
        \State Set $y' \gets y' + \{c_y\}$
    \EndFor
    \State \Return chunk-level altered summary $y'$
\EndFunction
\end{algorithmic}
\end{algorithm*}

\subsection{Ablation Studies}

We performed ablation studies on \textsc{DenoiseSum} by comparing it
to versions (a) using less synthetic training data, (b) using one kind
of noising function, and (c) with missing one module (explicit
denoising, partial copy, or discriminator).  We also compared our
model with a version that uses human-labeled categories, instead of
induced topic distribution, as the ground-truth category distribution
$p(z)$. For Rotten Tomatoes, we used movie genres (e.g., comedy,
drama) as categories, while for Yelp, we use business types
(e.g.,~restaurant, beauty parlor) as categories. In total, there are
21 movie genres and 898 business types.\footnote{The original versions
  of the datasets from \citet{wang2016neural} and
  \citet{chu2019meansum} do not contain this information. We share our
  version of these datasets here:
  \url{https://github.com/rktamplayo/DenoiseSum}.}

Table~\ref{tab:ablation_full} shows the ROUGE-1/2/L F1-scores of our
model and various versions thereof. The final model consistently
performs better on all metrics when compared to versions with less
synthetic data (second block), versions with only one type of noise
(second block), and versions with a module removed (third block). When
using human-labeled categories, we see a slight improvement in ROUGE-2
on Rotten Tomatoes, however the model performs substantially worse on
other metrics.  We believe there are several reasons for this.
Firstly, human-labeled categories, at least the ones available, are
not fine-grained enough to capture various aspects mentioned in the
reviews and their sentiment (e.g., did the actors perform well? was
the plot convoluted?).  Secondly, the number of business types
available on Yelp is very large (i.e., 898 types), which makes the
discriminator loss $\mathcal{L}_{disc}$ hard to optimize. This
explains the relatively larger decrease in performance on Yelp.

\begin{table*}[t]
  \small
  \centering
    \begin{tabular}{@{~~}lcccccc@{~~}}
    \thickhline
     & \multicolumn{3}{c}{Rotten Tomatoes}   & \multicolumn{3}{c}{Yelp} \\
    \multicolumn{1}{c}{Model} & ROUGE-1 & ROUGE-2 & ROUGE-L & ROUGE-1 & ROUGE-2 & ROUGE-L \\
    \thickhline
    \textsc{DenoiseSum} & \textbf{21.26} & 4.61 & \textbf{16.27} & \textbf{30.14} & \textbf{4.99} & \textbf{17.65} \\
    \hline
    \quad 10\% synthetic dataset & 20.16 & 3.14 & 15.39 & 28.54 & 3.63 & 16.22 \\
    \quad 50\% synthetic dataset & 20.76 & 3.91 & 15.76 & 29.16 & 4.40 & 17.54 \\
    \hline
    \quad no segment noising & 20.64 & 4.39 & 16.03 & 28.93 & 4.31 & 16.88 \\
    \quad no document noising & 21.23 & 4.38 & 16.22 & 28.75 & 4.06 & 16.67 \\
    \hline
    \quad no explicit denoising & 21.17 & 4.18 & 16.06 & 28.60 & 4.10 & 17.06 \\
    \quad no partial copy & 20.76 & 4.01 & 15.89 & 28.03 & 4.58 & 16.31 \\
    \quad no discriminator & 20.77 & 4.48  & 15.84 & 29.09 & 4.22  & 16.64 \\
    \quad using human categories &  20.67 & \textbf{4.69} & 15.87 &  28.54  & 4.02 & 15.86 \\
    \thickhline
    \end{tabular}%
    \caption{ROUGE-1/2/L F1 scores of our model and versions thereof with less synthetic data (second block), using only one noising method (third block), and without some modules (fourth block).}
  \label{tab:ablation_full}%
\end{table*}%

\subsection{Example Noisy Versions}

Figure \ref{fig:example_noising} shows example noisy versions of a
candidate summary using both segment and document noising
methods. Although segment noising yields texts which may not be
entirely comprehensible to humans, a few segments contain
understandable content that could be perceived as diverging
information and as such should not be included in the summary
(e.g.,~``some can't laugh hard'' in S.1 of Rotten
Tomatoes). Similarly, noisy versions generated by document noising
include content that is somewhat related but not critical for
generating the summary (e.g., ``remains just as relevant now as it did
in 1989'' in D.3 of Rotten Tomatoes).  These examples show that our
noise functions are not entirely random, in contrast to previous
trivial and un-informed approaches
\cite{fevry2018unsupervised,silberer2014learning}.

\begin{figure*}[t]
  \centering
\begin{small}
    \begin{tabular}{@{~~}llp{12cm}@{~~}}
      \thickhline
        Candidate Summary & &Quite possibly the greatest romantic comedy since some like it hot.\\
      \hline
      Segment Noise & S.1. & Some can't laugh hard of the best romantic comedy.\\
      & S.2. & The best romantic comedy set funny and unexpectedly moving, organically revealed but the sets something since the sexes. \\
      & S.3. & Movies created since its main pleasures' mystique recklessly assembled and all hers... love showcases of his stars.\\
      \hline
      Document Noise & D.1.  &Meg Ryan-Billy Crystal romantic comedy is hard not to like. \\
      & D.2. &MOV ... is an adult romantic comedy in a time when we don't get very many, and it has one thing going for it that gives it an enormous boost -- it's very funny. \\
      & D.3. & ... A better-than-average romantic comedy that remains just as relevant now as it did in 1989 \\      \thickhline 
\multicolumn{3}{c}{(a) Rotten Tomatoes} \\
%      \multicolumn{3}{c}{Yelp}\\
 %     \thickhline
      \multicolumn{3}{c}{} \\
      \multicolumn{3}{c}{} \\\thickhline 
      Candidate Summary & &Jeffrey in sales and Griffin in finance were awesome! we spent 2 long days at the dealership and these 2 were so patient with us trying to make up our minds. we believe we got a fair deal and would definitely go there again for our next vehicle. Thanks!\\
      \hline
      Segment Noise & S.1.  &I got the guys for Jeffrey there. We at least took a great deal, thanks proper expectations at would definitely go a Prius, and were the next morning. May of the same service advisor magic trick.\\
                          & S.2.  &We believe I but us with line! Check got Jeffrey awesome. our experience took for the bar, a real winner. Your trade would definitely go the autonation tempe website all employees having to block via.' Ve to drop very happy would recommend triple chocolate cupcake and red velvet cupcake pocket below a little - compared and the old owners name, dreamed the cupcake to bell again 1st time check would complain, back fixed a normal oil change and issued sure il 
                                    y a un grand choix. A new tire will never have start these 2 coming but our credit disrupted's disposition tries to throw good. To take our way, a hundred times have not been the next day. \\
                          & S.3. & So patient for all with a great deal! our experience. The guys how to pick 2 hours time is auto nation Toyota out voicemail, but the door again had the pleasure. My truck will definitely go there awesome, completely honest. You autonation spent my dash board from t for trick. \\
      \hline
      Document Noise & D.1. & Jason James was direct and honest. He gave a price that beat the competition, had the vehicle \& paperwork ready, and follower - up to make sure things were going well. Nicky, in finance, also did a good job. I hate car dealerships, but these guys did a good job and made it relatively painless.\\
                          & D.2.  &So me and my wife got a used van from the used car lot. The guy that helped us was pretty good but the used car manager is very rude and disrespectful. 3 weeks after we had the vehicle they told us they couldn't finance us. 3 weeks! when I went in to return the van the used car manager was talking on his office phone and typing on his personal cell phone and put his finger in order to tell me to wait. Finally after 10 minutes, another person took me over to finance to get my down payment back. They don't seem to know what they're doing over there. Would not recommend for future customers. \\
                          & D.3.  &Great service! I purchased a vehicle using an online  listing as I was located out of town from this dealership. The dealer (Jon Mefford) calmed any hesitation I had about purchasing a car from out of town and when I drove in to town to sign everything, the process was quick and seamless. Jon explained all the features of the vehicle and was very knowledgeable about any questions I had. Sofia, the person handling the finance part of things was also very personable and pleasant to deal with. Overall, a great experience!\\
      \thickhline
      \multicolumn{3}{c}{(b) Yelp dataset} \\
    \end{tabular}%
\end{small}
\caption{Noisy versions generated using segment and document noising
  on Rotten Tomatoes and Yelp.}
  \label{fig:example_noising}%
\end{figure*}%

\subsection{Human Evaluation on Amazon Mechanical Turk}

We conducted three different experiments using the Amazon Mechanical
Turk platform: the best-worst scaling evaluations for Rotten Tomatoes
and Yelp, and the summary veridicality experiment on the Yelp dataset.
For all experiments, we made sure crowdworkers had an approval rate of
98\% (or above) with at least 1000 tasks approved. Furthermore,
turkers were (self reported) native English speakers from one of the
following countries: Australia, Canada, Ireland, New Zealand, United
Kingdom, and United States.  We discuss further specifications for
each experiment in the next paragraphs.

\paragraph{Best-Worst Scaling}

For the best-worst scaling experiments, we created different templates
for each dataset. For Rotten Tomatoes, each Human Intelligence Task
(HIT) included the title of the movie and basic background
information: synopsis, release date, genre, director, and actors (see
the examples in Figure~\ref{fig:example_rotten}).  We also included
the human-written gold-standard summary (highlighted in blue),
emphasizing that the AMT workers must use it as a reference.  System
summaries were randomly shuffled and labeled A, B, or C.  Turkers were
then asked which to select the best or worst summary, according to
informativeness (i.e., does the summary present opinions about
specific aspects of the movie in a concise manner?), coherence (i.e.,
is the summary easy to read and does it follow a natural ordering of
facts?), and grammaticality (i.e., is the summary fluent and
grammatical?).  The criteria and their definitions were shown to
crowdworkers.  In total, 92~turkers participated in the study
annotating a total of 200~HITs.

For Yelp, each HIT also showed the name of the business and basic
information such as location and the type of service provided (see the
Figure~\ref{fig:example_yelp}.  Again, we showed the gold-standard
summary highlighted in blue, and randomly shuffled system summaries
(A, B, and C). Crowdworkers selected the best/worst summary according
to informativeness, coherence, and grammaticality. In total, 94
turkers participated in the study annotating a total of 200 HITs.

\paragraph{Summary Veridicality} In this experiment we only used the
Yelp dataset since the number of reviews is small and participants
could read them in a timely manner.  Each HIT presented the business
name, location, and type of service together with eight reviews and a
summary produced by one of the following systems: \textsc{MeanSum},
\textsc{DenoiseSum}, and \textsc{Gold}-standard summaries (see the
example in Figure~\ref{fig:example_yelp}).  Turkers were asked to
verify whether the facts mentioned in the summary were true (summaries
were at most ten sentences long). Specifically they had to decide for
each summary sentence whether it was fully supported, partially
supported, or not supported by the reviews. In total, 79 turkers
participated in this study annotating a total of 600 HITs.

\subsection{Example Summaries}

We show example summaries produced by three systems:
\textsc{SentiNeuron}, \textsc{MeanSum}, and our model
\textsc{DenoiseSum}, as well as the \textsc{Gold}-standard summary in
Figure \ref{fig:example_rotten} (for Rotten Tomatoes) and Figure
\ref{fig:example_yelp} (for Yelp). The extractive model
\textsc{SentiNeuron} tends to select reviews that are longer and more
verbose.  Summaries generated by \textsc{MeanSum} on Rotten Tomatoes
are mostly gibberish, which we argue is due to the model being unable
to handle the large number of input reviews in this dataset. Overall,
\textsc{DenoiseSum} produces the best summaries among the three
systems.

\begin{figure*}[t]
  \centering
\begin{small}
    \begin{tabular}{@{~~}lp{12cm}@{~~}}
      \thickhline
      \multicolumn{2}{c}{Movie: ``The Good Girl''} \\ 
      \hline
      Synopsis & Justine is thirty years old and works as a discount store clerk in Texas. Deeply unhappy in her marriage to a man who is infertile because of a dope-smoking habit, Justine soon begins an affair with Holden, the store's newly hired cashier and becomes pregnant. Holden, who has serious issues of his own, steals money from the store's safe for the two of them to run away, but the plan is short-lived when it takes a tragic turn for the worse. \\
      Released Date & Aug 7, 2002 \\
      Genre & Comedy, Drama \\
      Director & Miguel Arteta \\
      Actors & Jake Gyllenhaal as Holden, John C. Reilly as Phil, Tim Blake Nelson as Bubba, Jennifer Aniston as Justine, Zooey Deschanel as Cheryl \\
      \hline
      {\textsc{\textbf{Gold}}} & 
      A dark dramedy with exceptional performances from Jennifer Aniston and Jake Gyllenhaal, The Good Girl is a moving and astute look at the passions of two troubled souls in a small town. \\
      {\textsc{\textbf{SentiNeuron}}} & 
      Even during the most intense moments, it's hard to shake the impression that the conspicuously buff-and-polished Justine is only visiting this drab world, her miserable life an interesting career move. \\
      {\textsc{\textbf{MeanSum}}} & 
      Most of the time, the movie is a little too rare. \\
      {\textsc{\textbf{DenoiseSum}}} & 
      With good performances, direction and cast, The Good Girl is a provocative, intelligent and absorbing film. \\
      \thickhline
      \multicolumn{2}{c}{} \\ \thickhline
      \multicolumn{2}{c}{Movie: ``Iron Man 2''} \\ 
      \hline
      Synopsis & In "Iron Man 2," the world is aware that billionaire inventor Tony Stark is the armored Super Hero Iron Man. Under pressure from the government, the press and the public to share his technology with the military, Tony is unwilling to divulge the secrets behind the Iron Man armor because he fears the information will slip into the wrong hands. With Pepper Potts and James "Rhodey" Rhodes at his side, Tony forges new alliances and confronts powerful new forces. \\
      Released Date & May 7, 2010 \\
      Genre &  Action \& Adventure, Science Fiction \& Fantasy \\
      Director & Jon Favreau \\
      Actors & Robert Downey Jr. as Tony Stark, Gwyneth Paltrow as
               Virginia 'Pepper' Potts, Don Cheadle as Colonel James 'Rhodey' Rhodes, Mickey Rourke as Ivan Vanko/Whiplash, Sam Rockwell as Justin Hammer \\
      \hline
      {\textsc{\textbf{Gold}}} & 
      It isn't quite the breath of fresh air that Iron Man was, but this sequel comes close with solid performances and an action-packed plot. \\
      {\textsc{\textbf{SentiNeuron}}} & 
      Flabby, disjointed, and eschewing conflict for extended scenes of improv clowning, it's the superheroic equivalent of a rat pack film. \\
      {\textsc{\textbf{MeanSum}}} & 
      ... the movie has too many twists in its own way, but it's a bit too busy. \\
      {\textsc{\textbf{DenoiseSum}}} & 
      Iron Man 2 isn't as good as the first movie, but it is a fun and fascinating film. \\
      \thickhline
      \multicolumn{2}{c}{} \\ \thickhline
      \multicolumn{2}{c}{Movie: ``Sanctum''} \\  
      \hline
      Synopsis & The 3-D action-thriller Sanctum, from executive producer James Cameron, follows a team of underwater cave divers on a treacherous expedition to the largest, most beautiful and least accessible cave system on Earth. When a tropical storm forces them deep into the caverns, they must fight raging water, deadly terrain and creeping panic as they search for an unknown escape route to the sea. Master diver Frank McGuire (Richard Roxburgh) has explored the South Pacific's Esa-ala Caves for months. But when his exit is cut off in a flash flood, Frank's team--including 17-year-old son Josh (Rhys Wakefield) and financier Carl Hurley (Ioan Gruffudd)--are forced to radically alter plans. With dwindling supplies, the crew must navigate an underwater labyrinth to make it out. Soon, they are confronted with the unavoidable question: Can they survive, or will they be trapped forever? Shot on location off the Gold Coast in Queensland, Australia, Sanctum employs 3-D photography techniques Cameron developed to lens Avatar. Designed to operate in extreme environments, the technology used to shoot the action-thriller will bring audiences on a breathless journey across plunging cliffs and into the furthest reaches of our subterranean world. -- (C) Universal\\
      Released Date & Feb 4, 2011 \\
      Genre & Action \& Adventure, Drama, Mystery \& Suspense \\
      Director & Alister Grierson \\
      Actors & Richard Roxburgh as Frank McGuire, Ioan Gruffudd as Carl Hurley, Rhys Wakefield as Josh McGuire, Alice Parkinson as Victoria, Dan Wylie as Crazy George \\
      \hline
      {\textsc{\textbf{Gold}}} & 
      Sanctum is beautifully photographed, and it makes better use of 3-d technology than most, but that doesn't make up for its ham-handed script and lifeless cast. \\
      {\textsc{\textbf{SentiNeuron}}} & 
      In between the scary parts, we are subjected to a veritable Bartlett's of hackneyed dialogue. \\
      {\textsc{\textbf{MeanSum}}} & 
      You don't have to know the best thing about this film, it's the kind of movie that's not always a bad thing. \\
      {\textsc{\textbf{DenoiseSum}}} & 
      Sanctum isn't a great film, and it doesn't have a certain charm to overcome on top of its special effects. \\
      \thickhline
    \end{tabular}%
\end{small}
\caption{Examples of opinion summaries generated by three systems on
  the Rotten Tomatoes dataset.  We also show the human-generated
  consensus summary (\textsc{Gold}), as well as basic background
  information about the movie.}
  \label{fig:example_rotten}%
\end{figure*}%

\begin{figure*}[t]
  \centering
\begin{small}
    \begin{tabular}{@{~~}lp{12cm}@{~~}}
      \thickhline
      \multicolumn{2}{c}{Business: ``Noodle Pot''} \\ 
      \hline
      Location & Las Vegas \\
      Categories & Noodles, Specialty Food, Restaurants, Food, Ethnic Food, Chinese, Taiwanese \\
      \hline
      \multicolumn{2}{c}{Reviews} \\
      \multicolumn{2}{p{14.2cm}}{1. I thought this place okay. The Beef Roll was just average, not a big fan. The side dishes were okay ... cold cucumber, sliced tofu. I did however really like the wontons in red chili sauce.} \\
      \multicolumn{2}{p{14.2cm}}{2. Although I am not a big fan of beef noodles, I still wanted to come here and try their other dish. I ordered pig feet (i know, sounds scary) noodle soup.  I was not very impressed with it.  They were only 3 medium size pig feet in the soup, so it was not very filling.  I think the beef noodle soup would've been more filling.  But this place seems to be very popular as people kept coming in.  The turn over rate is pretty fast, so even if there is a line, I wouldn't think the wait would be too long.   If you are sick of buffets on the strip and feel like some hot (both temperature and spice) beef noodle soup, come and try this place.} \\
      \multicolumn{2}{p{14.2cm}}{3. The restaurant is really tiny and more of a cafe. The beef stew noodle is so perfect. Not too salty. Just enough beef, bok choy, and handmade noodles to satisfy any appetite. The pork chop noodle was also very good. It is a milder soup but the saltiness of the pork chop helps to balance it. The house-made chili gives everything the perfect kick and the burn pleasantly creeps up on you. The wontons in red sauce are also spectacular. I believe the red sauce has a vinegar base with hints of sweet and spicy. The restaurant is the perfect place to go during the winter. } \\
      \multicolumn{2}{p{14.2cm}}{4. "Nom nom nom nom nom. My mouth is watering. I think I'm due for a review too after eating here like everyday in November after discovering. My mom visited in March and we THOUGHT about going here but were sketchy about the cash only policy.  I have seriously missed out since then because I LOVE this place. The first time I came here my mom and I shared the beef noodle soup and the won ton soup. Both were so amazing with lots of fresh cilantro and green onion! It's def not a fancy place but this place has super great and flexible hours. I think they are open until 2am on Friday/Saturday. This is the perfect place to go on a chilly Vegas night!"} \\
      \multicolumn{2}{p{14.2cm}}{5. The best beef noodle soup that we had so far in the US. We live in California but never thought that we had to come to Vegas to find the beef noodle soup that we've been looking for!} \\
      \multicolumn{2}{p{14.2cm}}{6. Love this place...the best soup...I love those cucumbers...and very friendly stuff ...} \\
      \multicolumn{2}{p{14.2cm}}{7. "beef noodle soup and the pork/veggie dumplings are just what the doctor ordered. this 1-2 punch of savory goodness is gonna make your face royal-flushed when the sweet sweet msg kicks in. almost missed my flight out of vegas cuz of this meal but it was still worf it. l'oreal. the wifi password to this place is 12345678. is this real life?"} \\
      \multicolumn{2}{p{14.2cm}}{8. Zero star .......   They have really bad service... Wrong order, bad attitude, even worse, no apology for their mistake.....  And their food are horrible x 5....I will never ever come here again} \\
      \hline
      {\textsc{\textbf{Gold}}} & 
      This is a decent place to go for Chinese food. I really like the beef noodle soup. It has the perfect amount of salt. I wish the place wasn't so small and stuffy. I just didn't feel comfortable at all. Next time I'll just place a to go order. \\
      {\textsc{\textbf{SentiNeuron}}} & 
      The restaurant is really tiny and more of a cafe. The beef stew noodle is so perfect. Not too salty. Just enough beef, bok choy, and handmade noodles to satisfy any appetite. The pork chop noodle was also very good. It is a milder soup but the saltiness of the pork chop helps to balance it. The house-made chili gives everything the perfect kick and the burn pleasantly creeps up on you. The wontons in red sauce are also spectacular. I believe the red sauce has a vinegar base with hints of sweet and spicy. The restaurant is the perfect place to go during the winter. \\
      {\textsc{\textbf{MeanSum}}} & 
      This place has the best beef noodle soup. I had the chicken with broccoli and it was good. Its not spicy at all but not bad either. I have been coming here for over 10 years and have been back many times. They have a wide variety of dishes, not too oily, and good. The staff is friendly, clean, never disappointed. \\
      {\textsc{\textbf{DenoiseSum}}} & 
      I love this place. The restaurant is really clean and very friendly. The beef noodle soup is very good and the beef is so perfect. The food is very filling. I receive really good service here and recommend the place when you are in Vegas. The only complaint is that there's always a line every time I go here. I am definitely going back here again . \\
      \thickhline
    \end{tabular}%
\end{small}
\caption{Examples of opinion summaries generated by three systems on the Yelp dataset. 
  We also show the human-generated summary (\textsc{Gold}), as well as basic background information and the eight input reviews.}
  \label{fig:example_yelp}%
\end{figure*}%

\end{document}